\documentclass[letterpaper, 10 pt, conference]{ieeeconf}
\IEEEoverridecommandlockouts  
\usepackage{subcaption}
\usepackage{times}
\usepackage{multicol}
\usepackage[bookmarks=true]{hyperref}
\hypersetup{bookmarksopen,bookmarksnumbered,
pdfpagemode=UseOutlines,
colorlinks=true,
linkcolor=blue,
anchorcolor=blue,
citecolor=blue,
filecolor=blue,
menucolor=blue,
urlcolor=blue
}

\usepackage[utf8]{inputenc}
\usepackage{graphicx}
\usepackage{amsmath}
\usepackage{amssymb}
\usepackage{latexsym}
\usepackage{epsfig}
\usepackage{url}
\usepackage{booktabs}
\usepackage{multirow}
\usepackage{wrapfig}
\usepackage{xcolor}

\makeatletter
\let\NAT@parse\undefined
\makeatother

\usepackage[numbers]{natbib}

\newif\ifpaperfinal
\paperfinalfalse 
\ifpaperfinal
\newcommand{\tn}[1]{}
\newcommand{\ernote}[1]{}
\newcommand{\stnote}[1]{}
\newcommand{\vhnote}[1]{}
\newcommand{\ndnote}[1]{}
\else
\newcommand{\tn}[1]{\textcolor{teal}{\textbf{TN: #1}}}
\newcommand{\stnote}[1]{\textcolor{blue}{\textbf{ST: #1}}}
\newcommand{\ernote}[1]{\textcolor{red}{\textbf{ER: #1}}}
\newcommand{\vhnote}[1]{\textcolor{orange}{\textbf{VH: #1}}}
\newcommand{\ndnote}[1]{\textcolor{purple}{\textbf{ND: #1}}}
\fi

\title{\LARGE \bf Language-Conditioned Observation Models for Visual Object Search}

\author{Thao Nguyen, Vladislav Hrosinkov, Eric Rosen, Stefanie Tellex\\
\{thaonguyen, vladislav\_hrosinkov, eric\_rosen\}@brown.edu, stefie10@cs.brown.edu}

\begin{document}
%
%
%
%
%
%

\maketitle              

\begin{abstract}
  Object search is a challenging task because when given
  complex language descriptions (\textit{e.g.,} ``find the white cup on the table''), the robot must move its camera through the environment and recognize the described object.
  Previous works map language descriptions to a set of fixed object detectors with predetermined noise models, but these approaches are challenging to scale because new detectors need to be made for each object.  
  In this work, we bridge the gap in realistic object search by posing the search problem as a partially observable Markov decision process (POMDP) where the object detector and visual sensor noise in the observation model is determined by a single Deep Neural Network conditioned on complex language descriptions.
  We incorporate the neural network's outputs into our language-conditioned observation model (LCOM) to represent dynamically changing sensor noise.
  With an LCOM, any language description of an object can be used to generate an appropriate object detector and noise model, and training an LCOM only requires readily available supervised image-caption datasets. We empirically evaluate our method by comparing against a state-of-the-art object search algorithm in simulation, and demonstrate that planning with our observation model yields a significantly higher average task completion rate (from \textbf{0.46} to \textbf{0.66}) and more efficient and quicker object search 
  than with a fixed-noise model.
  We demonstrate our method on a Boston Dynamics Spot robot, enabling it to handle complex natural language object descriptions and efficiently find objects in a room-scale environment.

\end{abstract}

\section{Introduction}
Object search is a challenging task because the robot has incomplete knowledge of the environment, limited field of view, and noisy sensors. When asked to find an object in an environment, the robot must first infer the desired object from the language instruction, then efficiently move around the space to look for the object. Most images captured by the robot in this process will not contain the target object. Furthermore, even when the object is in the robot's field of view, it might not be detected due to occlusion, sensor noise, the viewing angle, the object's distance from the robot, etc.

\begin{figure}
  \centering
\subcaptionbox{A scene with the ground truth segmentation mask for the target object (``the green mug on the left").}{
  \includegraphics[width=0.22\linewidth]{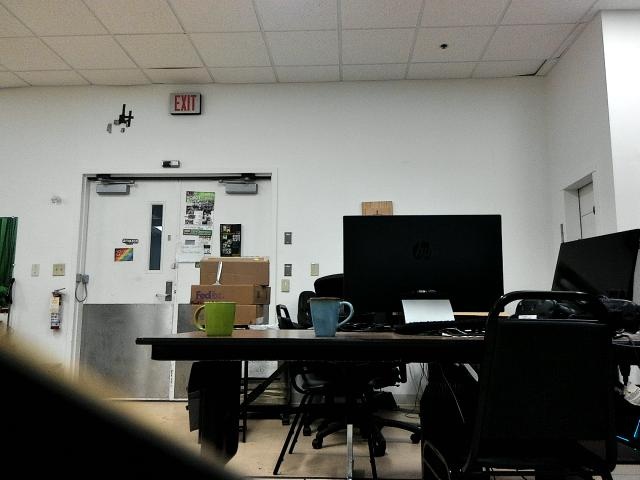}
  \includegraphics[width=0.22\linewidth]{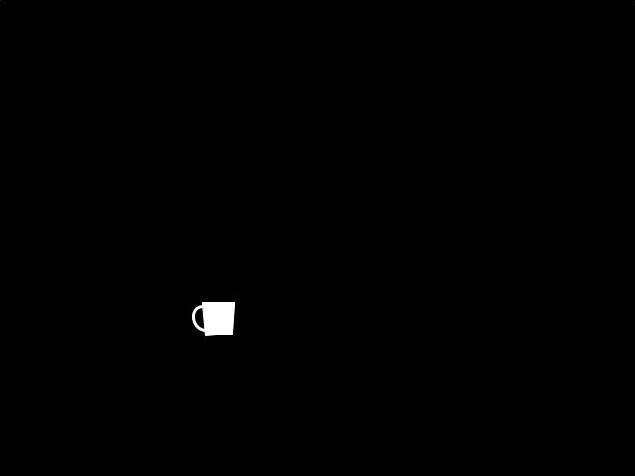}}\hfill
\subcaptionbox{Most images captured by the robot do not contain the target object.}{
  \includegraphics[width=0.22\linewidth]{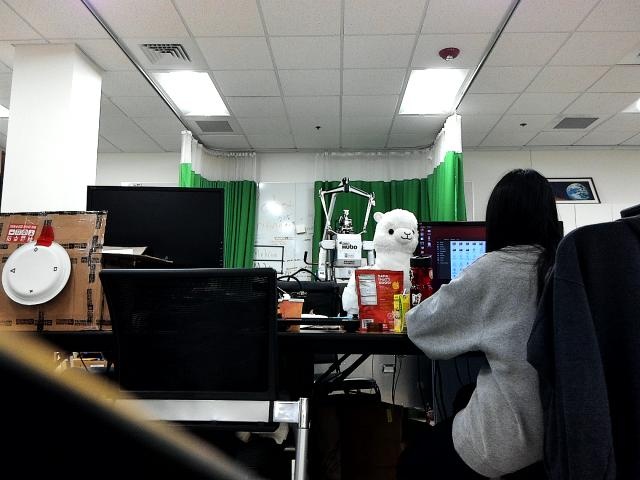}
  \includegraphics[width=0.22\linewidth]{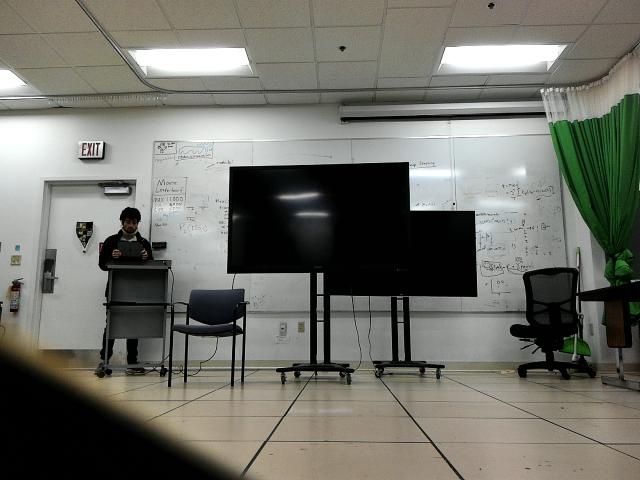}
  }
  \caption{Our system takes as input a natural language description of the target object, and constructs a detector for that object based on the description.  It addresses the problem that most images captured by a robot when searching for an object do not contain that object by incorporating a modified training process and using the confidence score of the detector in a POMDP model for object search.\label{fig:problem}}
\vspace{-5mm}
\end{figure}

There have been many previous works on improving the efficiency and accuracy of robot object search by using prior semantic knowledge \cite{kollar2009utilizing}, active visual search \cite{atanasov2013nonmyopic, aydemir2013active}, object manipulation \cite{li2016act, xiao2019online, danielczuk2019mechanical}, and belief factorization for multi-object search \cite{wandzel2019multi, zheng2020multi}. However, these works have often assumed that the target objects are specified with very simple language (such as ``cup'' for the object's class), and thus cannot fully utilize more complex language descriptions (such as ``white cup'') to avoid exhaustively searching over similar object instances in the environment.
Furthermore, the robot is usually assumed to have a fixed-accuracy object detector \cite{kollar2009utilizing, aydemir2013active, wandzel2019multi, zheng2020multi} or that detection noise only comes from occlusion \cite{li2016act, xiao2019online}. This is challenging to scale as new occlusion models and detectors have to be made for new objects.
Additionally,
modeling the detector as having a fixed accuracy prevents the robot from dynamically reasoning about observations it gets from the detector. This means the robot is unable to decide to gather more data instead of trusting a low-confidence detection, or trust a high-confidence detection more easily, potentially leading to reduced object search success rates and efficiencies.
The computer vision community has developed deep learning models that can detect objects with high accuracy \cite{he2017mask, redmon2016you}, even given complex language descriptions of the desired object such as ``white cup on the left" \cite{hu2016segmentation,gu2021open,kuo2022findit}. However, these models often assume that the object is somewhere in the input image and must be localized within that image.  In contrast, when searching for objects, most images captured by the robot will not contain the object being searched for~(Figure~\ref{fig:problem}).  


Our work addresses these problems by embedding a deep-learned object
detector within a Partially Observable Markov
Decision Process (POMDP) \cite{cassandra1998survey}. Our approach takes as input a language description of an object, and uses it to condition a camera-based observation model that is used to plan object-search actions and update the agent's belief about the object's pose. To achieve this, we modify the training process for the detector from
\citet{hu2016segmentation} to handle the large number of images which do not contain the target object, and incorporate the detector's confidence scores into the POMDP belief updates.
This allows us to handle complex language descriptions and
search for objects more efficiently in household environments by reasoning dynamically.
Our contributions are five-fold:

\textbf{1)} An experimental analysis on confidence scores outputted from language-conditioned visual segmentation models as a proxy for different sources of observation noise.

\textbf{2)} A novel class of visual observation models (Language-Conditioned Observation Models -- LCOMs) whose detections and parameters are conditioned on natural language.

\textbf{3)} A novel decision making framework for object search that leverages LCOMs to use natural language to account for scene-dependent detection accuracy when estimating state uncertainty for planning.

\textbf{4)} A set of experiments on simulated robot hardware that compare the performance of planning models using LCOMs against fixed-noise sensor models on the object search task.

\textbf{5)} A demonstration of our method on a Boston Dynamics Spot robot \cite{spot}, which enables Spot to handle complex natural language object descriptions and efficiently find objects in a room-scale environment, without using fiducial markers.

\section{Related Work}

Related work for robot object search generally falls into one of two
categories: \textit{model-based} and \textit{end-to-end policy
  learning}. Model-based approaches separate state estimation and
planning to leverage probabilistic inference, whereas
model-free approaches leverage deep neural networks to learn a policy end-to-end.

There is a collection of works that employ deep learning for
end-to-end visual and object search \cite{faust2018prm,
  wortsman2019learning, niroui2019deep,gadre2022clip} or modular differentiable components \cite{chaplot2020object,majumdar2022zson}. Our work differs from these
in that we perform model-based planning to leverage our known dynamics
models.  Model-based planning has the potential to generalize better
to new environments and systems with less training data because we
encode a model of the robot's sensor and actuation capabilities, and only use deep learning for visual processing. 

POMDPs \cite{kaelbling1998planning} are a framework for sequential decision
making under uncertainty frequently used for object search
problems. \citet{li2016act} and \citet{xiao2019online} treat object
search in clutter as a POMDP that can be efficiently solved by using
approximate online planners and constraining planning samples based on
spatial constraints and conditioning action selection on the current
belief state, respectively. However, their observation models are only
based on occlusion statistics calculated from object region
overlap. Our proposed observation model can instead account for errors
not solely derived from occlusion by conditioning on complex
language. \citet{danielczuk2019mechanical} train a deep learning model
to segment colored masks for objects in a pile from RGB-D images and
score each mask on whether it belongs to the target object. They, however, use a fixed object priority policy for action selection and assume a fixed sensor pose, while we focus on planning how to explore
a space for the purpose of object search by leveraging an active
sensor.

\citet{aydemir2013active} frame the object search problem as active visual search and calculate candidate viewpoints based on a probability distribution over the search region, which is informed by prior knowledge of correspondences between objects and semantic room categories. However, they do not account for sensor error and assume the object to be detected if it is in the robot's field of view.
\citet{atanasov2013nonmyopic} plan a sequence of sensor views to effectively estimate an object's orientation. These approaches are similar to our work in that they account for realistic sensor model errors, but unlike our work they do not use a general camera-based object detector. 

\citet{wandzel2019multi} introduce Object-Oriented POMDP (OO-POMDP) to factorize the robot's belief into independent object distributions, enabling the size of the belief to scale linearly in the number of objects, and employ it for efficient multi-object search.
\citet{zheng2020multi} extend OO-POMDP for efficient multi-object search in 3D space. Both of these works assume simple language inputs and fixed-accuracy object detectors. Our work
builds on these frameworks but
instead explores using a deep-learned detector that takes as input a natural language phrase and camera image to create an object detector that models varying levels of accuracy.

The computer vision community has developed deep learning models trained on object segmentation datasets that can detect objects with high accuracy \cite{he2017mask, redmon2016you, hu2016segmentation,gu2021open,kuo2022findit}. The models self-supervisedly learned to output confidence scores along with their detection results.  The output confidence scores do a good job of reflecting the models' detection accuracy, which dynamically changes depending on the input images. We build on the model developed by \citet{hu2016segmentation} for our object detector because it is trained to handle referring expressions---complex noun phrases to describe objects.


\section{Preliminaries}
POMDPs are a framework for modeling sequential decision making problems where the environment is not fully observable.
Formally, a POMDP can be defined as a tuple $<S, A, \Omega, T, O, R>$, where $S, A, \Omega$ denote the state, action, and observation spaces of the problem, respectively. After the agent takes action $a \in A$, the environment state transitions from $s \in S$ to $s' \in S$ following the transitional probability distribution $T(s, a, s') = p(s'|s, a)$. As a result of its action and the state transition, the agent receives an observation $z \in \Omega$ following the observational probability distribution $O(s', a, z) = p(z|s', a)$, and a real-valued reward $R(s,a)$.

Because the environment is partially observable, the agent does not have full knowledge of the current state $s$ and instead maintains a \emph{belief state} $b$ which is a probability distribution over the states in $S$. The agent starts with an initial belief $b_0$ and updates its belief after taking an action and receiving an observation (where $\eta$ is the normalizing constant): $b'(s') = \eta O(s', a, z) \sum_{s \in S} T(s, a, s') b(s)$.

A policy $\pi$ is a mapping from belief states to actions. The agent's task is to find a policy that maximizes the expected sum of discounted rewards given an initial belief:\\
$V^{\pi}(b_0) = \mathbb{E} \left[ \sum_{t=0}^{\infty} \gamma^t R(s_t, a_t) \bigg| a_t=\pi(b_t) \right]$
where the discount factor $\gamma$ determines the impact of future rewards on current decision making.

While many problems can be modeled by POMDPs, they are typically computationally
intractable for exact planning in large domains \cite{madani1999undecidability}. To address the planning complexity, we use the sampling-based planner PO-UCT \cite{silver2010monte}, which uses Monte-Carlo tree search with upper confidence bounds to select an action by estimating the best state-action values using rollouts conditioned on states sampled from the current belief state, and then performs an exact belief update each time step based on the incoming observation and performed action. PO-UCT has successfully been used in robotic object-search settings \cite{wandzel2019multi,zheng2020multi}. However, we note that our contribution is not dependent on the specific method of planning and state estimation, and LCOMs would be useful in any approach that requires a model of the observation probability distribution.


\section{Object Search Formulation}
We model object search as a POMDP with an observation model corresponding to a deep-learned object detector.

\subsection{Planning Framework}
\label{sec:planning}
To model the object search problem, we assume access to an occupancy-grid map, $M$, which is an $m \times n$ grid that marks locations as either empty or occupied, and is used for defining the space of positions and directions in the environment. We assume an object is completely contained within one of the grid cells in the map.  Our main contribution is the novel language-conditioned observation model (LCOM), which modifies the observation model dynamically based on the results of the deep-learned object detector, and which we describe in detail in Section \ref{sec:obs_model}. Formally, we define the object search POMDP problem as a 10 tuple: $<o_{d},L,S,A,T,R,\gamma,\Omega,h_{L},O>$

\begin{enumerate}

    \item $o_{d}$: is a desired object that exists in the environment (not including the robot). The desired object $o_{d}$ has a 2D position attribute $(x_{o_{d}},y_{o_{d}}) = o_{d}$, representing its discrete position in the occupancy-grid map $M$. The desired object is used to define the reward function.
    
    \item $L$: is a string of words representing the natural language command given by the human, such as ``The white cup on the table.'' 
    $L$ is only used to condition the visual observation model and transform raw images into our fan-shaped sensor model. We defer more details to Section \ref{sec:obs_model}. In this work we assume $L$ to be given at the start and remain constant throughout the interaction, and defer handling dynamical language to future work.
    
    \item $S$: is a set of states, where each state $s \in S$ is a $2$ dimensional vector $(o_{d},r) = s$, where $r=(r_{x},r_{y},r_{o})$ is a 2D position and discrete orientation (NORTH, EAST, SOUTH, WEST) for the robot in $M$. We assume $r$ is fully observable and $o_{d}$ is only partially observable, yielding a mixed-observable state. This assumption is equivalent to assuming our robot is equipped with a LIDAR sensor and has previously run SLAM \cite{pritsker1984introduction} and localized itself within that map, but does not know where the desired object is currently located.
    
    \item $A$: is a set of actions the robot can execute to move around the map, observe object locations, and declare the desired object as found. Specifically, we have three types of parameterized actions:
    \begin{enumerate}
        \item $Move$(\textit{DIR}): points the robot in direction \textit{DIR} and moves it one grid in that direction, with \textit{DIR} being either NORTH, EAST, SOUTH, WEST.
        \item $Look$: has the robot execute a look action from its current position and orientation $(r_{x},r_{y},r_o)$.
        \item $Find(X,Y)$: has the robot attempt to find the desired object $o_{d}$ at grid cell $(X,Y)$. If $o_{d}$ is at $(X,Y)$, the action will mark the object as found and terminate the episode.
    \end{enumerate}
    
    \item $T$: is a deterministic transition function, where $Move$ actions transition the robot to different states by changing its position and orientation $(r_{x},r_{y},r_o)$. $Find$ actions can transition the robot to a terminal state after finding the desired object.  
    
    \item $R$: is a reward function, where all $Move$ actions receive $-2$ reward each, $Look$ receives $-1$ reward, and $Find(X,Y)$ receives a $1000$ reward when done at the location of the desired object $(X,Y) == (x_{o_{d}},y_{o_{d}})$ and $-1000$ otherwise. 
    
    \item $\gamma$: is the discount factor, which we set to $0.9$.
    
    \item $\Omega$: is the set of observations from our sensor, where each $\omega \in \Omega$ is a pair of RGB and depth images.
    
    \item $h_{L}$: $\Omega \rightarrow {z^{s},c^{s}}$ is a language-conditioned observation-mapping function that transforms raw images into observations from the same fan-shaped sensor model described in \citet{wandzel2019multi} and confidence scores for each object detection. 
    If the desired object $o_{d}$ is not detected by the sensor, $z^{s} = NULL$. Otherwise, $z^{s}$ is the location $(X,Y)$ where $o_{d}$ is detected in the discretized fan-shaped region $V$.
    $c^{s}$ represents the object-specific observation confidence score.
    We describe how $h_{L}$ is used for LCOMs in Section \ref{sec:obs_model}, and our particular instantiation of $h_{L}$ for our experiments in Section \ref{sec:detector}.
    
    \item $O$: is the Language-Conditioned Observation Model (LCOM), which assigns probabilities to observations $z_{t}$ based on the current state $s_{t}$, action $a_t$, and natural language command $L$. The $Move$ actions always produce the $NULL$ observation, the $Look$ action produces noisy fan-shaped measurements conditioned on the language, and the $Find(X,Y)$ action always produces the $NULL$ observation except when $(X,Y) == (x_{o_{d}},y_{o_{d}})$. We discuss the observation model in more detail in the following subsection.
\end{enumerate}

\begin{figure}
    \centering
    \includegraphics[width=0.9\linewidth]{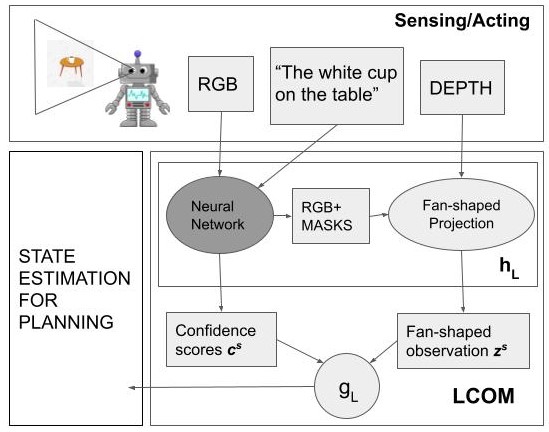}
    \caption{\textbf{LCOM Overview}: The robot receives an RGB-D image and language description of the object. RGB-D and language go into $h_{L}$, which produces language-conditioned confidence scores $c^s$ for our fan-shaped detected observations $z^s$. The confidence scores are then transformed by $g_{L}$ into a noise model for the detected observations, which is used to update the belief about the object's location via state estimation. Ovals are algorithms, and rectangles are data. The shaded oval is learned.
    }
    \label{fig:overview}
\vspace{-5mm}
\end{figure}

\subsection{Language-Conditioned Observation Model (LCOM)}
\label{sec:obs_model}
Figure \ref{fig:overview} presents an overview of LCOM.
When we receive an RGB-D sensor observation, $\omega$, we can transform it into our fan-shaped sensor observation $z^{s}$ and associated confidence scores $c^{s}$ by using the language-conditioned observation mapper $h_{L}(\omega) = z^{s} ,c^{s}$. LCOMs are independent of any particular instantiation of $h_{L}$ as long as they satisfy the functional definition described in Section \ref{sec:planning}, and for the rest of this section's discussion we treat $h_{L}$ as a black box function. In our experiments, we instantiate $h_{L}$ using a deep neural network.

We treat $z^{s}$ as having a probability of being drawn from three mutually exclusive and exhaustive events: a true positive ($A$), a false positive ($B$), or a true or false negative ($C$). More formally, let $A$ be when $z^{s}$ is from the desired object $o_{d}$ and $z^{s} \in V$, $B$ be when $z^{s} \in V$ but $z^{s}$ comes from other sources besides $o_{d}$, and $C$ be when $z^{s} = NULL$. We assume the $Find(X,Y)$ action always give perfect information about the potential object at location $(X,Y)$ (\textit{i.e.,} observations resulting from $Find$ are not language-conditioned). In simulation this is reflected by knowing the ground truth state, and in real-life this can be reflected by asking a human to verify the selected location. For the $Look$ action, we parameterize the probability of each of the events and the noise model for the observation conditioned on that event based on the associated confidence score $c^{s}$,
and decompose the observation model $p(z^{s}|s,a)$ into:
\begin{equation}
    p(z^{s}|s,a,c^{s}) = \sum_{e \in \{A, B, C\}} p(z^{s}|e, s,a,c^{s}) p(e|s,a,c^{s})
\end{equation}
If event $A$ occurs, the observation is distributed normally with $\mu$ being the true object position: $p(z^{s}|A,s,a,c^{s}) = \eta'Norm(z^{s}|\mu,\Sigma)$. $\eta'$ is the normalization factor, and the covariance matrix is defined by $\Sigma=\sigma \textbf{I}^{2\times2}$.
If event $B$ occurs, the observation is distributed uniformly within the sensor region: $p(z^{s}|B,s,a,c^{s}) = \frac{1}{|V|}$. If event $C$ occurs, the null observation has nearly 1 probability while any other observation has nearly 0 probability, which we implement with additive smoothing.

Similar to \citet{wandzel2019multi}, we define the probability of the events as $p(A|s,a) =\alpha$, $p(B|s,a) = \beta$, $p(C|s,a) = \gamma$, where $\alpha + \beta + \gamma = 1$. The probability of these events are conditioned on whether or not the desired object $o_{d}$ is in the fan-shaped sensing region $V$, which is defined as:
\begin{equation}
(\alpha, \beta, \gamma) =  \begin{cases} 
      (\epsilon_{TPR}, 0, 1-\epsilon_{TPR}) & \textnormal{if $o_{d}$ is in $V$} \\
      (0, 1-\epsilon_{TNR}, \epsilon_{TNR}) & \textnormal{if $o_{d}$ is not in $V$}
   \end{cases}
\label{eq:epsilon}
\end{equation}
where $\epsilon_{TPR}$ represents the sensor's true positive rate, and $\epsilon_{TNR}$ represents its true negative rate.



Together $\sigma, \epsilon_{TPR}$, and $\epsilon_{TNR}$ define the sensor's overall accuracy. 
To implement the function $g_{L}$, which transforms the confidence scores to the sensor noise in the observation model, we map the continuous value of $c^{s}$ to a discrete range of hyper-parameter values that represent high-confidence and low-confidence for each setting, respectively. In our experiments, we map  $\epsilon_{TPR}$ to $0.7$ and $\sigma$ to 0.6 when $c^{s} \geq 1$, and  $\epsilon_{TPR}$ to $0.5$ and $\sigma$ to 1 otherwise. These numbers reflect that when the detector's confidence is high, the true positive rate should be high and the uncertainty over the observed position of the object should be low.
We note that LCOMs depend on visual input to detect potential objects in the image and report confidence scores that are used to define the sensor noise in the observation model. During state estimation with real-robot hardware, acquiring visual input is straightforwardly done by capturing images with the robot's camera. During planning, however, acquiring visual input may be challenging because it requires synthesizing novel images based on the pose of the robot and potential location of the target object. For computational efficiency, when performing visual object search in our experiments, we only use LCOMs for updating the agent's belief during state estimation, and use a fixed observation model similar to \citet{wandzel2019multi} during planning based on the 2D geometries of the known occupancy-grid map $M$. Integrating different 3D scene representations into the planning module is straightforward but orthogonal to our contribution, so we defer this investigation to future work.



\subsection{Object Detector}
\label{sec:detector}
We build upon the model developed by \citet{hu2016segmentation} for our object detector as it can handle complex noun phrases to describe objects. The model takes in a referring expression and RGB image and outputs scores for every pixel in the image, which are then binarized and returned as the predicted segmentation mask for the image region described by the language. The loss function used for training is the average pixelwise loss: $Loss = \frac{1}{WH} \sum_{i=1}^{W} \sum_{j=1}^{H} L(v_{ij},M_{ij})$. $W$ and $H$ are image width and height, $v_{ij}$ is the pixel's score, and $M_{ij}$ is the binary ground-truth label at pixel $(i,j)$.

The original model by \citet{hu2016segmentation} was trained on the ReferIt dataset \cite{kazemzadeh2014referitgame} which mostly contains outdoor images, whereas we are interested in detecting indoor household objects. We, therefore, additionally trained the model on the RefCOCO dataset \cite{kazemzadeh2014referitgame} which contains referring expression annotations for segmented objects from the COCO dataset of common objects in context \cite{lin2014microsoft}.
Furthermore, the original model was primarily trained on positive examples such that most images contained the target object, and the model only had to learn to identify where the object was in the image.  In contrast, when using a model like this for object search, most images fed to the model will not contain the referenced object. Thus filtering a large number of true negatives without missing the rare true positive is key to good performance in search tasks. We augmented the model's training data with negative examples where the object described by the referring expression does not appear in the image, and thus the model should return an empty segmentation mask.
Our model, trained on the augmented data with a learning rate of 0.01, achieved a true negative rate of \textbf{0.918}, a significant improvement over the original model's true negative rate of \textbf{0.124}.

We now describe our instantiation of $h_L$ for our experiments based on the deep learning segmentation model. The model takes in the RGB image and language $L$ and outputs a segmentation mask---a binary image which identifies pixels that are part of the target object described by $L$.  If the mask is empty, the model did not detect the object and $z^s=NULL$. Otherwise, we take the average of the depth value at each pixel in the mask as well as the coordinates of the mask's center point and project it
into a location $(X,Y)$ in the robot's fan-shaped sensing region (\textit{i.e.,} fan-shaped projection) and return $(X,Y)$ as $z^s$. We also retain the model's original output score for each pixel, which we average over all pixels in the mask and use as the confidence value $c^s$ for the detection. We note that the scores were not specifically trained for this task.

\begin{figure}
    \centering
    \includegraphics[width=\linewidth]{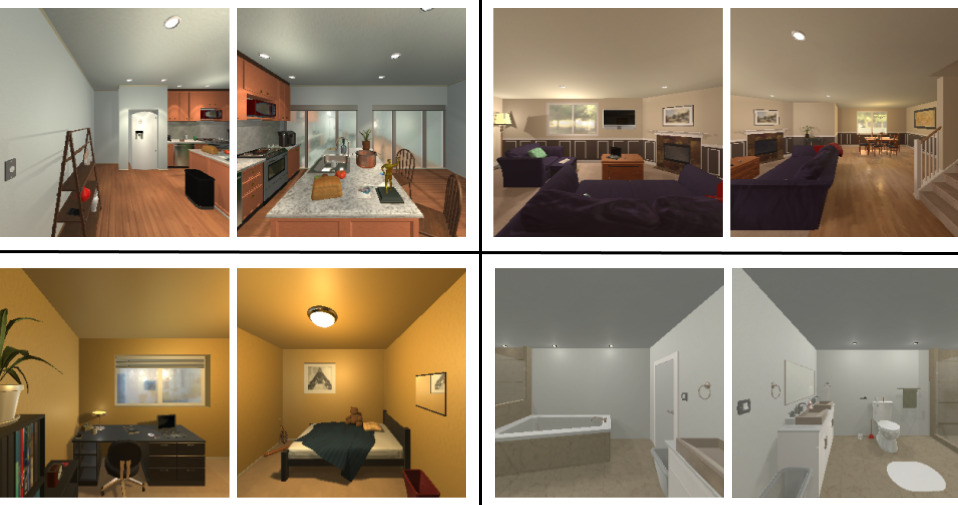}
    \caption{\textbf{Simulated Scenes}: example images of the AI2-THOR scenes used in our experiments. The scene categories are: kitchen \textit{(top left)}, living room \textit{(top right)}, bedroom \textit{(bottom left)}, and bathroom \textit{(bottom right)}.}
    \label{fig:rooms}
\vspace{-5mm}
\end{figure}

\section{Experiments and Results}
Our aim is to test the hypothesis that language-conditioned observation models combined with POMDPs can increase a robot's speed and accuracy in finding objects in complex environments.  We evaluated our system both in a variety of simulation environments and on a real physical robot.

\subsection{Simulation Results}
\label{sec:sim}
We use scenes from the AI2-THOR simulator \cite{ai2thor} to conduct our experiments. AI2-THOR consists of 120 near photo-realistic 3D scenes spanning four different categories: kitchen, living room, bedroom, bathroom. We select a subset of 15 scenes with 30 target objects (for an average of 2 objects/scene) for our experiments. Figure \ref{fig:rooms} shows images of the scenes used in our experiments. The average size of a scene is $4 \times 4$ meters, which we discretize into a $16 \times 16$ cell grid map with each cell being $0.25 \times 0.25$ meters.


We build upon the POMDP implementation by \citet{zheng2020pomdp_py} in the pomdp\_py library. We modeled the POMDP as having no prior knowledge of the target object's location, thus it had a uniform initial belief state over all possible object locations. We used a planning depth of $3$, exploration constant of $10000$, planning time of $10$ seconds for each action, and gave the agent a maximum time of $5$ minutes and $10$ $Find$ actions to complete each object search task. We generated simple natural language descriptions of the objects in our experiments as input to the agent.

\begin{figure}[t]
  \centering
  \subcaptionbox{}{
    \includegraphics[width=0.5\linewidth]{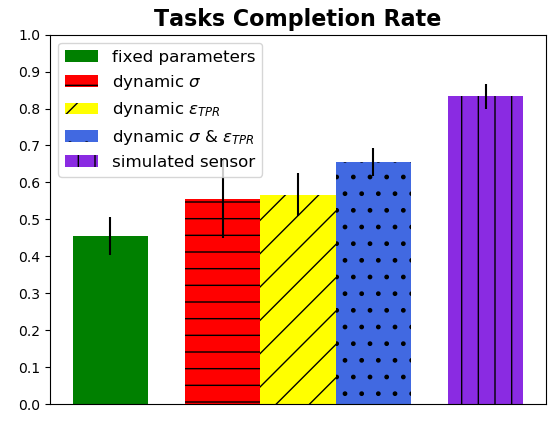}}%
  \subcaptionbox{}{
\includegraphics[width=0.5\linewidth]{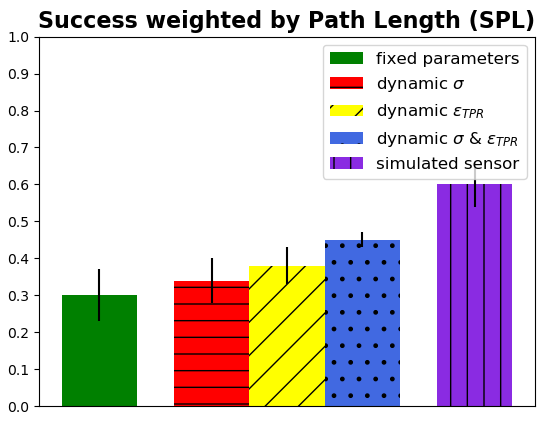}}
\vspace{-2mm}
\caption{\textbf{Simulation Results}: Task completion rates and success weighted by path lengths for a simulated sensor and deep-learned sensor with static/dynamic observation models.}
\label{fig:graph}
\vspace{-6mm}
\end{figure}

Results appear in Figure~\ref{fig:graph}.  We present both the task
completion rate---the percentage of time the robot successfully finds
the object, and success weighted by normalized inverse path length (SPL) \cite{anderson2018evaluation}. SPL is calculated as: $\frac{1}{N}\sum_{i=1}^{N}S_i\frac{l_i}{max(p_i,l_i)}$
where $N$ is the total number of tasks, $l_i$ is the shortest path from the agent to the goal for task $i$, $p_i$ is the path the agent actually took for the task, and $S_i$ is a binary indicator of success in the task. For our experiments, $p_i$ is the number of actions the agent actually took to search for the object, and $l_i$ is the lowest number of actions needed to find the object. If the agent achieves a higher task completion rate but took more steps overall to find the objects, it will have a lower increase in its SPL. We collected $l_i$ by performing planning with a perfect sensor with no noise. The perfect sensor was able to find all 30 objects at an average of 7.8 actions per object search task.

Each different version of our model was tested 3 times and we report the average and standard error in their performance. We present results for fixed optimal values of the sensor parameters computed from the scenes in our dataset. Our deep learning model achieved a true positive rate (TPR) of 0.581, a true negative rate (TNR) of 0.918, and a covariance of 0.827 for the normal distribution over the desired object's position.
We then show results with $\sigma$, $\epsilon_{TPR}$, and both $\sigma$ \& $\epsilon_{TPR}$ values set dynamically based on the deep-learned detector's output confidence score
As the sensor's TNR is already high, we decide to keep $\epsilon_{TNR}$ fixed. Lastly, we show the performance with a simulated sensor whose noise model perfectly matches the model used for planning by the POMDP.

As expected, the performance for the simulated sensor is the best.  This
is because the sensor observations are being generated from ground truth with exact noise models.  This provides an upper bound on our system's performance, and also indicates that if we used a more realistic sensor model, our system has the potential to perform even better.  In particular, the simulated sensor will sample multiple images with the same viewpoint independently, which is not true for the deep-learned detector. All versions of our system with a dynamic
observation model outperform the static version. In addition, the version with dynamical $\sigma$ \& $\epsilon_{TPR}$ achieved a significantly higher average task completion rate and SPL than the static version (from \textbf{0.46} to \textbf{0.66}, and from \textbf{0.30} to \textbf{0.45}, respectively). On average, this version took \textbf{10.1} actions and \textbf{104} seconds to find each object, compared to the \textbf{11.5} actions and \textbf{118} seconds taken by the static version.
Overall, these results demonstrate that using a dynamic observation model significantly improves the ability
of our system to find objects quickly and efficiently in realistic environments.

\begin{figure}
    \centering
    \includegraphics[width=0.4\linewidth]{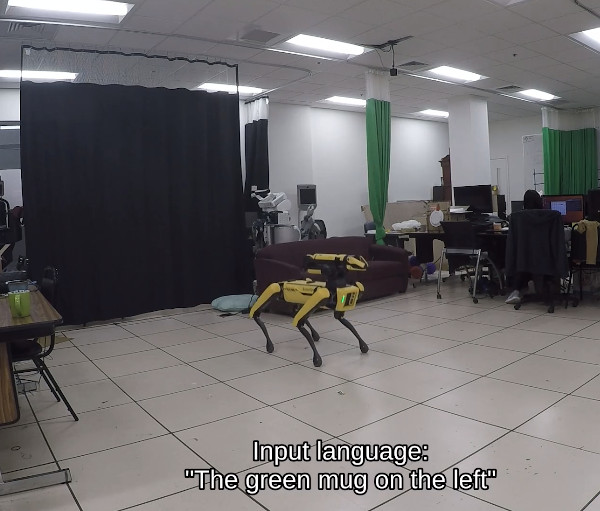}
    \includegraphics[width=0.43\linewidth]{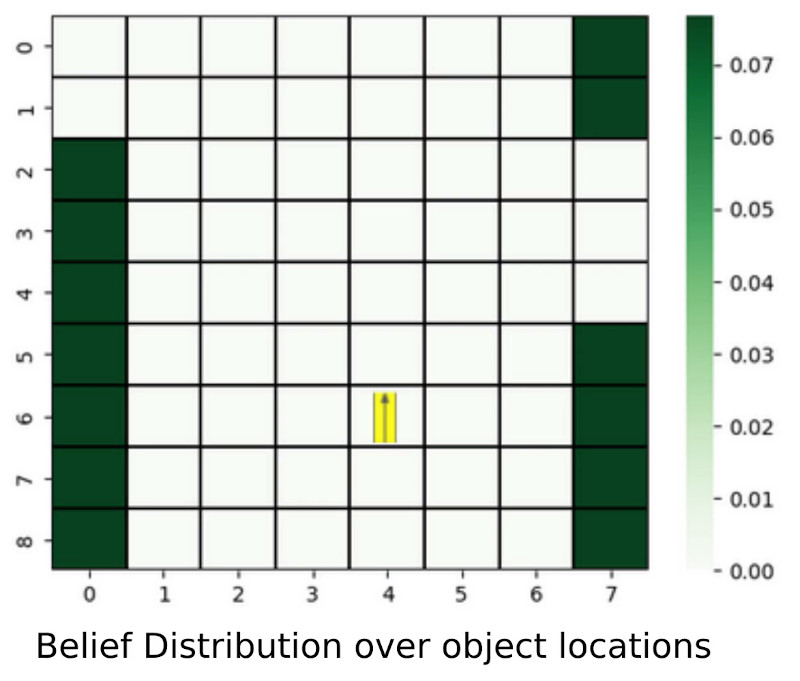}
    \vspace{-2mm}
    \includegraphics[width=0.43\linewidth]{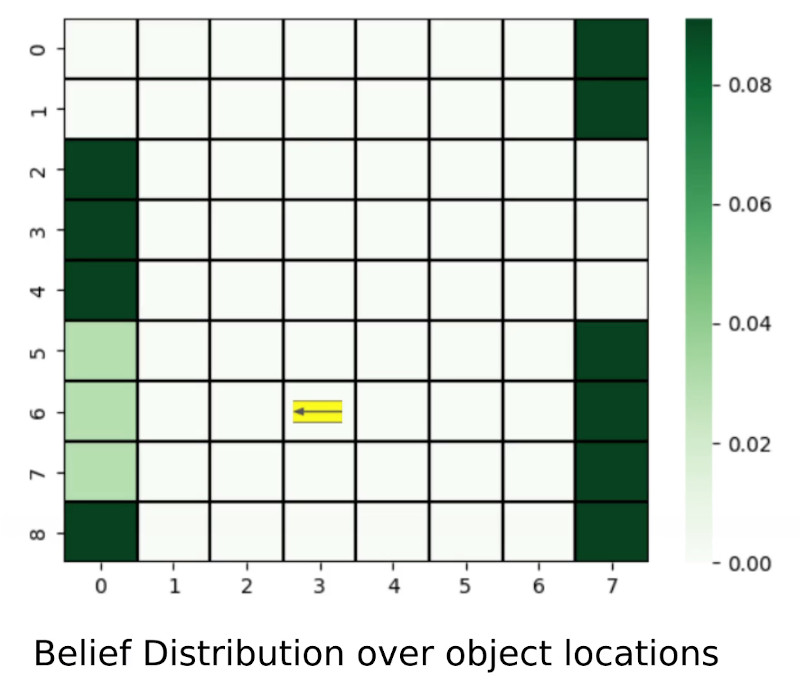}
    \includegraphics[width=0.42\linewidth]{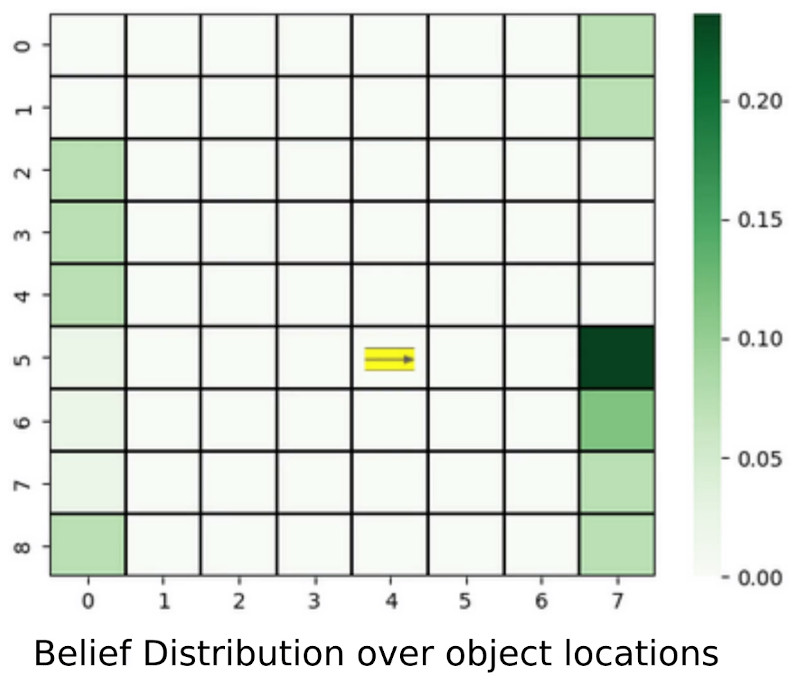}
    \caption{\textbf{Real Robot Demonstration}: sample images from our real robot experiments with the Spot using LCOMs to find an object. \textit{top left:} the robot is turned on and tasked with finding ``the green mug on the left.'' \textit{top right:} the robot's uniform initial belief about the target object's location. \textit{bottom left:} the robot moves and looks at a part of the room where the object is not located, and updates its belief that the object is most likely somewhere else. \textit{bottom right:} the robot moves and looks where the object is actually located, and after updating its belief has maximum likelihood estimate at the target object's true location.}
    \label{fig:robot}
\vspace{-6mm}
\end{figure}

\subsection{Real-World Demonstration}

We provide a real-world demonstration on the Boston Dynamics Spot robot.  The robot takes as input an occupancy-grid map of the environment
and a typed natural language phrase describing the desired object. RGB and Depth images are taken from two separate cameras in the robot's hand, and pixel correspondence between the two images is computed using both cameras' intrinsic and extrinsic matrices.
Spot moves through the environment by taking steps that are $0.6$ meters (one grid cell) in length, and all decisions are driven by the POMDP until it finds the object. Scenes from our demonstration and the corresponding belief updates from using LCOMs with real robot hardware are shown in
Figure~\ref{fig:robot}. Full video footage of the robot executing the task, the incoming sensor data, and the LCOM outputs is available at \url{https://youtu.be/3Z4XQUQXCsY}. The robot was asked to find ``the green mug on the left" and successfully did so in $8$ actions, where the planning and execution of each action took $10$ seconds. This demonstration shows our system runs on a real-world platform in a realistically sized environment, computes a
policy and observations efficiently, and enables a robot to
efficiently search for and find objects.

\subsection{ViLD Experiments}
\begin{figure}[]
  \centering
  \subcaptionbox{}{
    \includegraphics[width=0.5\linewidth]{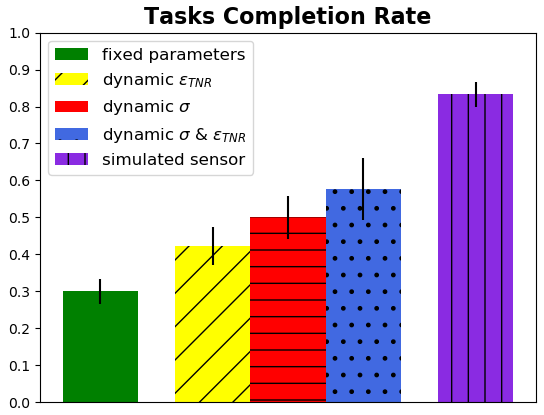}}%
  \subcaptionbox{}{
\includegraphics[width=0.5\linewidth]{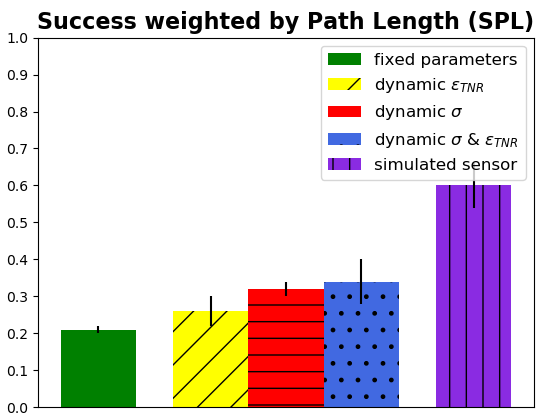}}
\vspace{-2mm}
\caption{\textbf{ViLD Simulation Results}: Task completion rates and success weighted by path lengths for a simulated sensor and ViLD with static/dynamic observation models.}
\label{fig:clip}
\vspace{-6mm}
\end{figure}

Given the recent success in open-vocabulary image classification/object detection powered by CLIP \cite{radford2021learning}, we swapped out our trained object detector with ViLD \cite{gu2021open}, an object detector trained via vision and language knowledge distillation, for our object search experiments. ViLD takes in natural language expressions and an RGB image, proposes regions of interest within the image, computes visual embeddings for the regions, and calculates the dot product (score) between the visual embeddings and text embeddings generated by CLIP. It then returns the highest scoring image regions that correspond to the input natural language.

For each object in our experiment, we pass a simple natural language description of the object into ViLD along with the RGB image taken by the robot, and take as output the segmentation mask associated with the highest scoring image region.
The mask has the same size as the input RGB image, and value 1 for every pixel within the proposed region and 0 otherwise. If ViLD does not find an image region corresponding to the input language, the segmentation mask is empty and the observation $z^s=NULL$. Otherwise, we take the average of the depth value at each pixel in the mask as well as the coordinates of the mask's center point and project it into a location $(X,Y)$ in the robot's fan-shaped sensing region and return $(X,Y)$ as $z^s$.
We also retain the image region's score as the confidence score $c^{s}$.

Without fine-tuning, ViLD achieved a true positive rate (TPR) of 0.976, a true negative rate (TNR) of 0.118, and a covariance of 1.825 for the normal distribution over the desired object's position on our AI2-THOR dataset. Similar to other object segmentation methods, ViLD tends to return a non-empty segmentation mask even when the queried object is not in the input image.

Experiment results are shown in Figure \ref{fig:clip}. The settings are the same as those described in Section \ref{sec:sim}. We present results for fixed values of the ViLD sensor parameters. Next, we show results with $\sigma$, $\epsilon_{TNR}$, and both $\sigma$ \& $\epsilon_{TNR}$ values set dynamically based on the output confidence score $c^s$. We map  $\epsilon_{TNR}$ to $0.1$ and $\sigma$ to $1.0$ when $c^{s} \geq 0.25$, and  $\epsilon_{TNR}$ to $0.3$ and $\sigma$ to $2.0$ otherwise. As ViLD's TPR is already high, we decide to keep $\epsilon_{TPR}$ fixed. Each different version was tested 3 times and we report the average and standard error in their performance.

The simulated sensor's performance is still the upper bound on the object search task.
ViLD's performance trails behind our object detector which is fine-tuned for the task. However, all versions of our system with a dynamic observation model still significantly outperform the static version. The version with dynamical $\sigma$ \& $\epsilon_{TNR}$ achieved an average task completion rate of \textbf{0.578} and SPL of \textbf{0.34}, compared to the \textbf{0.3} and \textbf{0.21} achieved by the static version. These results demonstrate that our system works seamlessly with different object detectors and using dynamic observation models improves object search performance.

\section{Conclusion}
Our contribution is a novel observation model that uses the detector's confidence score to better model the detection accuracy. This enables us to handle complex language descriptions of objects and perform object search with a real object detector in realistic environments. In addition, our method can be easily adapted to new environments without having to relearn the observation model's parameters.

Our model only considers 2D space.  In future work, we plan to extend
to 3D models, building on \citet{zhengmulti21} and \citet{fang2020move} to model the 3D
structure of objects.  This extension will enable the robot to reason about different 3D viewpoints and predict the structure of a
partially observed object to gather more views to identify
and localize it. We also plan to specifically train the detector's output confidence scores to represent its detection accuracy.

Additionally, our model cannot reason about the
likelihood of different views of the same object to improve
its detection/localization of that object.  Our current observation
model assumes that each observation is independent, so if the robot observes the same scene from the same viewpoint, it
will become more and more certain whether the object is present or
not.  However, in practice, when viewing an image
from the same viewpoint, a deep-learned detector will give the same
results; the observations are not independent samples.  In the future,
we could address this problem by creating a new observation model based on inverse graphics and an expected 3D model of the object appearance, enabling the robot to predict the next best view to maximally reduce its uncertainty about the object's location.

Furthermore, we focus on language descriptions of the desired object to generate the object detector and observations. More complex language instructions that provide information about the location of the object such as ``look in the kitchen'' or ``the object is to your right" can be incorporated by directly updating the agent's belief about the object's pose.

Overall we see object search as a central problem for human-robot
interaction, as finding, localizing, and then grasping an object is a
first step for almost anything a person wants the robot to do in the
physical world.  Embedding object search as a sub-component of a more
sophisticated dialog system can enable the robot to engage in
collaborative dialog with a human partner to interpret complex natural
language commands, find and manipulate objects being referenced, and
fluidly collaborate with a person to meet their needs.


\section*{Acknowledgments}
The authors would like to thank Nick DeMarinis for all his support and help. This work was supported by NSF under grant awards IIS-1652561 and CNS-2038897, AFOSR under grant award FA9550-21-1-0214, and Echo Labs.

\bibliographystyle{IEEEtranN}
\bibliography{references}

\end{document}